# Multimodal Machine Learning in Precision Health

Adrienne Kline[1], Hanyin Wang[1], Yikuan Li[1], Saya Dennis[1], Meghan Hutch[1], Zhenxing Xu[2], Fei Wang[2], Feixiong Cheng[3] and Yuan Luo[1]*

1. Department of Preventative Medicine, Northwestern University, Chicago, 60201, IL, USA
2. Department of Population Health Sciences, Cornell University, New York, 10065, NY, USA
3. Cleveland Clinic Lerner College of Medicine, Cleveland Clinic, Cleveland, 44195, OH, USA

*Correspondence to: Dr. Yuan Luo, Department of Preventative Medicine, Northwestern University, Chicago, 60611, U.S.A., yuan.luo@northwestern.edu

## Summary

**Background:** As machine learning and artificial intelligence are more frequently being leveraged to tackle problems in the health sector, there has been increased interest in utilizing them in clinical decision-support. This has historically been the case in single modal data such as electronic health record data. Attempts to improve prediction and resemble the multimodal nature of clinical expert decision-making this has been met in the computational field of machine learning by a fusion of disparate data. This review was conducted to summarize this field and identify topics ripe for future research.
**Methods:** We conducted this review in accordance with the PRISMA (Preferred Reporting Items for Systematic reviews and Meta-Analyses) extension for Scoping Reviews to characterize multi-modal data fusion in health. We used a combination of content analysis and literature searches to establish search strings and databases of PubMed, Google Scholar, and IEEEXplore from 2011 to 2021. A final set of 125 articles were included in the analysis.
**Findings:** The most common health areas utilizing multi-modal methods were neurology and oncology. However, there exist a wide breadth of current applications. The most common form of information fusion was early fusion. Notably, there was an improvement in predictive performance performing heterogeneous data fusion. Lacking from the papers were clear clinical deployment strategies and pursuit of FDA-approved tools.
**Interpretation:** These findings provide a map of the current literature on multimodal data fusion as applied to health diagnosis/prognosis problems. Very few papers compared the outputs of the multimodal approach with unimodal predictions. However, those that did achieved an average increase of 4.2% in predictive accuracy (AUC). Multi-modal machine learning, while more robust in its estimations over unimodal methods, has drawbacks in its scalability and the time-consuming nature of information concatenation.
**Funding:** The study is supported in part by NIH GrantsU01TR003528 and R01LM013337.

## Introduction

Clinical decision support has long been an aim for those implementing algorithms and machine learning in the health sphere[1-3]. Examples of algorithmic decision supports utilize lab test values, imaging protocols or clinical (physical exam scores) hallmarks.[4,5] Some health diagnoses can be made on a single lab and a single threshold. This is the case in diabetes in older adults[6]. Other diagnoses made based on a constellation of the signs, symptoms, lab values and/or supportive imaging are referred to as a 'clinical diagnosis'. Oftentimes these clinical diagnoses are based on additive scoring systems that representing a threshold that requires crossing, or a mixture of positive and negative hallmarks required prior to confirmatory labeling. One example of such a diagnostic protocol is with Parkinson's disease. Two supportive features are required to be included in the diagnosis such as bradykinesia, rigidity, rest tremor or a beneficial response to dopaminergic therapy. Exclusionary criteria for Parkinson's include but is not limited to; a history of repeated strokes with an associated stepwise Parkinsonian manifestation, or severe autonomic failure (e.g. orthostatic hypotension)[7]. This example highlights the expected benefit of leveraging both imaging data (for history of strokes), combined with structured Electronic Health Records (EHR) that provides blood pressure measurements for indicative of orthostatic hypotension and clinical exam findings.

The modus operandi of a clinical diagnosis may fail to consider the importance of relative weighting of these disparate data inputs due to the limitations of human decision-making capacity and potentially non-linear relationships. The strength of algorithmic decision-making support is that it can be used to offload such tasks, ideally yielding a more successful result. This phenomenon has sparked an interest in fusion studies using health care data. The purpose of this study is to highlight the current scope of this research domain.

> **Mutlimodal machine learning**: the area of machine learning concerned with bringing together disparate data sources to capitalize on the unique and complimentary information in an algorithmic framework
> **Data harmonization**: the improvement of data quality and utilization through the use of machine learning, through interpretation of existing characteristics of data and action taken on data to suggest subsequent data quality improvements
> **Mutliview machine learning**: another eponym for multimodal machine learning
> **Data fusion**: Refers to the specific methodology undertaken to perform data integration for mutlimodal/multiview machine learning. These come in 3 broad categories; early, intermediate/joint, and late

Undertakings to characterize this literature have been performed by Huang et al.[8], who performed a systematic review of deep learning fusion of imaging and EHR data in health. However, it was limited to EHR and imaging data, where only deep learning applications were shown. In a follow-up paper[9] that included commentary on omics and imaging data fusion. The current study is more inclusive in the breath of the types of machine learning protocols used and attempts to encompass all current modalities.

Data fusion is underpinned by information theory and is the mechanism by which disparate data sources are merged to create an information state based on the sources' complementarity[10]. Data fusion is concisely defined as: ". . . the process of combining data to refine state estimates and predictions."[11] updating the commonly used definition: 'A process dealing with the association, correlation, and combination of data and information from single and multiple sources to achieve refined position and identity estimates, and complete and timely assessments of situations and threats, and their significance.'[12].

The hope and expectation in machine learning is that doing so will result in an improvement in predictive power[13,14] and therefore more reliable results in potentially low validity settings[15].

Data fusion touts the advantage that the results of modeling become inherently more robust, by relying on a multitude of informational factors rather than a single type. However, the methodology
of combinatory information has drawbacks; adding complexity to specifying the model and reducing the interpretability of results[15,16].

Successful data fusion is built on the pillar of data harmonization, that is quality control of data performed alongside the models in which it will be implemented. Because data from different sources and file formats are rarely uniform, and this holds true especially for clinical data[17]. For example, data sets can have different naming conventions, units of measure, represent different local population biases. This feat can be difficult, and a balance is required between efficacious harmonization (information that is similar and works together) and pure (information that corresponds 1:1)[18]. Care must be taken to search and correct for systematic differences between datasets and assess the degree of interchangeability. For example, Colubri et al. in order to aggregate computed tomography (CT) and PCR labs, they performed an intra-site normalization to ensure that values were comparable across sites and needed to discard several potentially informative clinical variables since they were not available in all datasets[19]. The clinical field of Heart Failure with preserved ejection fraction (HFpEF) saw novel applications of multiple tensor factorization formulations to integrate the deep phenotypic and trans-omic information[20], and this extends to other areas of precision medicine[21]. To increase the portability of EHR-based phenotype algorithms, the Electronic Medical Records and Genomics (eMERGE) network has adopted common data models (CDMs) and standardized design patterns of the phenotype algorithm logic to integrate EHR data with genomic data and enable generalizability and scalability[22-25].

Given the recency of this literature, there is no consensus on the optimal way to combine data. Classification of informational source amalgamation was first posed by Durrant-Whyte[26]. Incoming data can be complementary; where sources represent different aspects of the entity under investigation, redundant; where sources provide the same information and serve to reinforce the information, and lastly cooperative; where the information is fused to create new information of a higher order. It is this last class or cooperative information that data fusion takes its roots.

There are three main types of data fusion that are used in machine learning; early (data-level), intermediate (joint), and late (decision-level)[27]. In the case of early fusion, multiple data sources are converted to the same information space. This often results in vectorization or numerical conversion from an alternative state, as performed by Chen et al. via vectorized pathology reports[28]. Images possess characteristics that can undergo numerical conversion based on area, volume, and/or structural calculations[29]. These are then concatenated with additional measurements from structured data sources and fed into an individual classifier. Canonical correlation analysis[30], non-negative matrix factorization[31,32], Independent Component Analysis (ICA) and numerical feature conversion methodologies exist as common options to trans-form all data into the same feature space[33].

Intermediate data fusion occurs inside a step-wise set of models and offers the greatest latitude in model architecture. A 3-stage deep neural learning and fusion model was proposed by Zhou et al.[34] where stage 1 consists of feature selection by a soft-max classifier for independent modalities. Stage 2 and 3 constitute combining these selected features, establishing a further refined set of features, and feeding these into a Cox-nnet to perform joint latent feature representation for Alzheimer's diagnosis. In contrast to early fusion, intermediate fusion combines the features that distinguish each type of data to produce a new representation that is more expressive than the separate representations from which it arose.

In late fusion, multiple models are trained, where each model corresponds to an incoming data source. This is akin to ensemble learning, which offer better performance over individual models[35]. Ensemble methods use multiple learning algorithms (typically applied to the same dataset) to obtain better predictive performance than could be obtained from any of the constituent learning algorithms alone. However, in the case of multimodal ML ensemble here can refer to ensemble learning within a data type and across data types. Common ensemble learning methods include bagging, boosting, stacking, random forest, and Bayesian optimal classifier. These takes symbolic representations as sources and combine them to obtain a more accurate decision[36]. Bayesian's methods are typically employed at this level[26] to support a voting process between the set of models into a global decision. A broad-based schematic for the 3 subtypes of data fusion are presented in Figure 1.

Early and late fusion have the greatest flexibility as to the number of models that can be fused. Early fusion gains this advantage by converting all data into the same feature space that can be classified using simpler models such as support-vector machines or logistic regression. While there inevitably exists a loss of information via dimensionality reduction using this method, it involves a single model and therefore offers a propitious entry point for those wishing to perform fusion for health predictions. Attribute differences and similarities are showcased in Table 1.

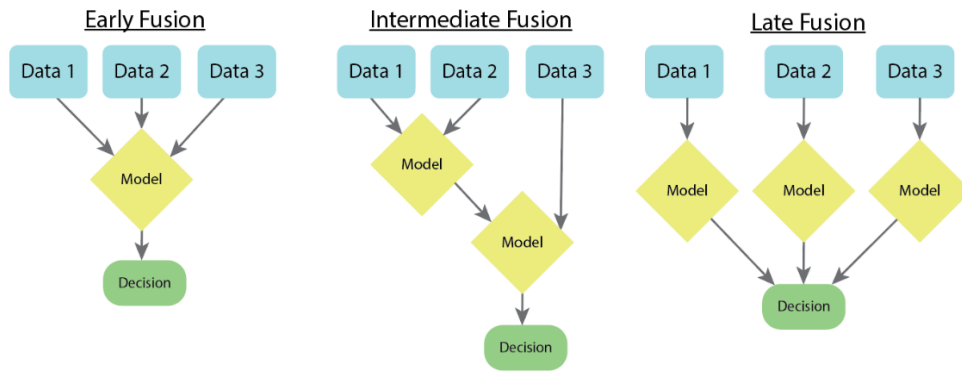

*Figure 1:* **early, intermediate, and late fusion data flow and general structure**

| Attribute | Early | Intermediate/Joint | Late/Decision |
|---|---|---|---|
| Ability to handle missing data | no | no | yes |
| Scalable | no | yes | yes |
| Multiple models needed | no | yes | yes |
| Improved accuracy | yes | yes | yes |
| Voting/weighting issues | no | yes | yes |
| Interaction effects across sources | yes | yes | no |
| Interpretable | yes | no | no |
| Implemented in health | yes | yes | yes |

*Table 1.* **Comparison of Fusion techniques**

## Methods

There are several research questions that drove this review:

***RQ1****- What characterizes the published literature using multimodal data fusion in the health sector?*
A multiplicity of different lenses will be used to showcase these studies.

***RQ2****- What are the different analysis techniques, methods, and strategies applied to analyze multi-modal health data for diagnosis/prognosis?*
In addition to providing a summary of the analytical techniques applied, this RQ aims to explore the challenges and opportunities that researchers have encountered in implementing these techniques.

***RQ3****- What areas of heterogeneous data fusion have had the most impact?*
This RQ aims to identify current gaps in the literature that will provide recommendations for future information concatenation in the health sector. Also, how the results of this RQ allow different health care researchers the opportunity to make informed decisions on how to use multi-modal data fusion as part of their studies.

Search Strategy and Selection Criteria

Inclusion requirements are: an original research article published within the last 10 years which inclusively encompassed years 2011-2021, published in English and on the topic of multi-modal or multi-view using machine learning in health for diagnostic or prognostication applications. 'multi-modal' or 'multi-view' for our context and sake of brevity is taken to mean data sources not of the same type. For example, a paper using CT and MRI would be considered multi-modal imaging but under our premise was considered to be uni-modal i.e. imaging. Exclusions for these purposes of this review were scientific articles not published in English, commentaries or editorials, other review articles. Papers were also excluded if the data was not human derived. We also excluded papers where the fusion already occurred at the data generation stage, such as spatial transcriptomics producing integrated tissue imaging and transcriptomics data[37-39]. All papers underwent a 2-person verification for inclusion in the manuscript.

Search strings were established via literature searches, domain expertise. Additional keywords were identified based on key word co-occurrence matrices established from the abstracts of the previous articles found. Table 2 displays the search strings, where an individual string would include one keyword from each column. For example, "`health + heterogeneous data + machine learning`" would be one of the search strings for automatic search. This process was repeated until all combinations of search strings were incorporated. An overview of the inclusion/exclusion process is noted in Figure 2 and echoes the standard set by PRISMA extension for scoping reviews[40]. All titles included in this review can be seen in Supplementary material.

| Keyword$_1$ | Keyword$_2$ | Keyword$_3$ |
|---|---|---|
| health \| medicine | heterogeneous \| fusion \| multi-modal \| multi-view | learning [machine, deep] \| artificial intelligence |

*Table 2.* **health-related keyword, Multimodal-related keyword, machine learning-related keywords, | : or**

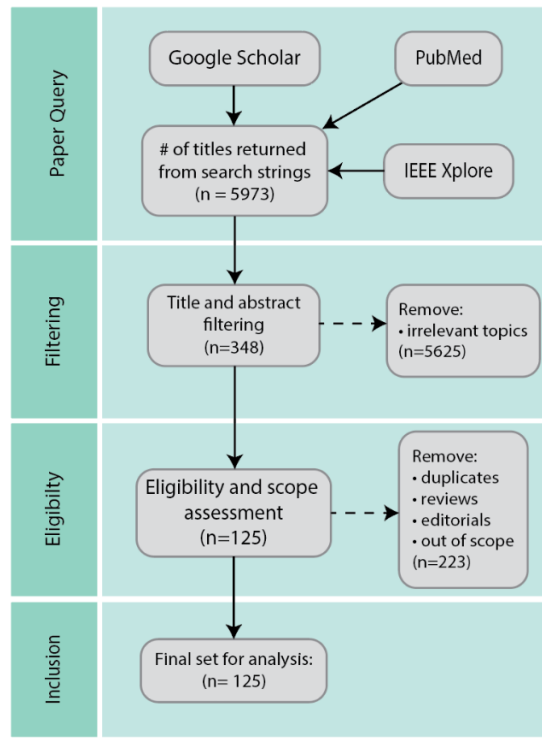

*Figure 2:* **Overview of Study Inclusion Process**

Data Analysis

Information garnered from the articles included title, year published, FDA approval of the tool, whether published in a clinical journal, author affiliations, number of authors, locations (continents) and abstract. Health context information extracted included the specific health disease under investigation and the healthcare domain(s) that this fall under. It should be noted that several topics fell into one or more broad based health topics. For example, the specific health care topic might be small-cell lung cancer. The corresponding health care domains would be respiratory and cancer.

Given the multi-modality aspect of machine learning, we also recorded and extracted the number of different modalities used and the divisions (i.e. text/image vs EHR/genomic/time series). To elucidate this, a paper characterized using lab values (structured EHR) and computed tomography as input into machine learning would be categorized as 'Imaging/EHR' subtype. The objective of each paper was extracted in a 1-2 sentence summary along with the keyword (if available). Patient characterization in the studies noted was performed by ascertaining the number of unique patients leveraged in the cohort and patient sex (i.e. Men/women/both or not mentioned).

Computational information extracted via recording the coding interface(s) used in data processing/analysis, machine learning type, data merging technique (early, intermediate, late) and types of machine learning algorithms used. Validation of the results of any machine learning work is important. Statistical tests run, whether validation was performed (yes/no) and the nature of that validation and other outcomes measures were all recorded for each paper. Validation consisted of leave-one-out cross validation, train-test split, n-fold cross validation. Statistical tests comparing outcomes from varying models included but were not limited to: student's t-test, Chi-square, ANOVA etc. Outcome measures included accuracy, F1, recall, sensitivity, specificity etc. Assessing the significance, impact and limitations of each paper was extracted by noting their primary findings and individual limitations as noted in the papers.

**Results**

The topic modelling displayed in Figure 3 showcases the category, specific health ailment under investigation and the modality type. Neurology and singularly Alzheimer's had the most papers published on this topic, accounting for 22 of the publications[33,41]. Individual health diseases/diagnoses are mapped to all applicable overarching health topic areas as noted previously in the data extraction section above. With the advent of the COVID-19 pandemic, several primary research articles are dedicated (5)[42-46], which can be arrived at through respiratory or infectious disease hierarchies. Articles are subsequently mapped back to their multi-modal subtype on the right of the Sankey plot.

This should serve as a resource to fellow researchers to identify areas that lack such saturation, namely dermatology[47], hematology[48], medication/drug issues such as alcohol use disorder investigated by Kinreich et al. that may offer new research horizons[49]. A dashboard resource published in conjunction with this review article is available here https://multimodal-ml-health.herokuapp.com/. This dashboard was created as an interactive infographic-based display of some major findings presented here. To assist and spawn future work, a drop-down menu has been created to help researchers filter our underlying data file of titles based on overarching health topic by selection, so they may more readily find and read the papers as presented whose broader trends are presented here.

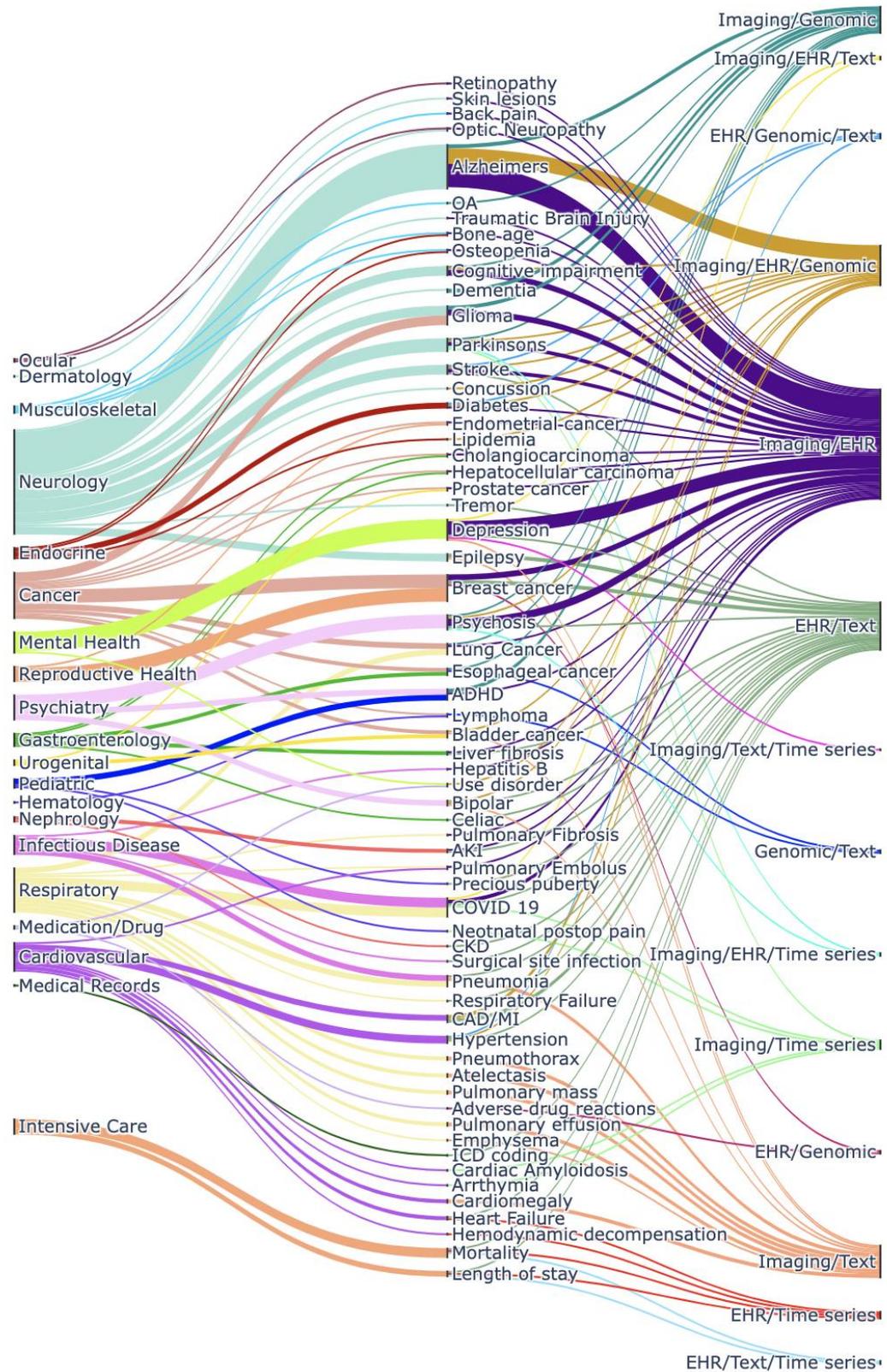

*Figure 3:* **Topic and Modality Modeling**

Of the models used in the papers, 128/130 explicitly reported performing a validation procedure of their model(s). The most common validation performed were; N-fold cross validation (54)[50,51], train test split (55), leave one out cross validation (9) and external dataset (10). A cornucopia of machine learning techniques and methods were comprised both within and across articles in this review. They have been summarized in Table 3 noting which fusion umbrella subtype they were implemented in.

| Fusion Type | Machine learning models/techniques implemented |
| --- | --- |
| Early | SVM[28-31,33,49,50,52-80], RF[28,30,33,66,68-70,72,73,77-79,81-88], Gaussian process[33,54], Bayesian network[28,89], NB[53,70,72-74,79], n-grams[53], LR[28,46,50,66-68,78,79,88,90-94], ridge regression[76,86], multivariate linear regression[51,76,86,87], K-means[74], DT [28,31,67,68,74,75,78,83,92,95], MLP[68,80,87,96,97], transfer learning[47,61,93,98,99], AutoEncoders[56,95,100,101], bag-of-words[80], Transformers (BERT, GPT)[99], RNN [96,102], LASSO[86,90], CNN[47,79,80,93,99,100,102,103], GNN[101,104], Semantic-embeddings[102], DNN[60], multitask learning[61], gradient boosting classifiers [70,71,73,92], Markov model[105], bagging[87], ensemble learning[64,87], KNN[64,72,74,75,106], ANN[72,77], Adaboost[67,72,74,78], tensor decomposition[73], SGD[86], MARS[86], MKL[63,107], cox regression[108], DUN[109], EM[106], iMSF[106], mixture model[110], graph clustering[105], network/graphical Lasso[105], hierarchical clustering[105] |
| Intermediate | LR[111-115], SVM [34,58,111,113,114,116-121], DT[112,122], CNN[41,44,114-119,121-135], AutoEncoders[41,134], multivariate regression[58,111,120,130,136], DNN[34], n-grams[114], MLP[115,121,124,129,130,135], LSTM[112,124,126,128-130,136], RF[44,48,58,112,114,115,118,120,134], Semantic-embeddings[122], LASSO[122,137], NB[58], bi-LSTM[58,122], ensemble learning[120], gradient boosting[137], transfer learning[128], multitask learning[136,138], linear regression[118], XGBoost[115,121], Adaboost[115], ANN[118], word2vec[132], bag-of-words[121], NLP[132,139], ridge regression[137,140], SVR[133], MKL[48,133], LDA[139], graph learning[139] |
| Late | SVM[141-146], RF[41,43,143-146], KNN[43,147], CNN[41,43,145,147-152], LASSO[146,153], MLP[36,42,45], LSTM[36,148], AutoEncoders[36,41], DT[147,152], LR[45,143,145,146], bi-LSTM[42,152], mixed effects linear model[154], ensemble learning[155], multivariate linear regression[144], DNN[143], NLP[42,147], XGBoost[42,146,156], NB[147], sentiment analysis[42], word embedding[45,147], word2vec[45], GLMNET[156], recursive partitioning[156], GAM[156], graph clustering[157] |

SVM: support vector machine, RF: random forest, LR: Logistic regression, DT: decision trees, CNN: convolutional neural network, GNN: graph neural network, NB: naive Bayes, KNN: k-nearest neighbors, MKL: Multiple Kernel Learning, DUN: deep unified networks, SGD: stochastic gradient descent, MARS: Multivariate adaptive regression splines, MLP: multilayer perceptron, GLM- NET: Elastic-Net Regularized Generalized Linear Models, NLP: natural language processing, bi-LSTM: bidirectional long short-term memory, DNN: deep neural network, LDA: Latent Dirichlet Allocation, SVR: support vector regression, GAM: generalized additive model, EM: Expectation Maximization, iMSF: incomplete Multi-Source Feature

*Table 3*. **Fusion and machine learning methods included in this review**

All papers noted in this review used either two or three disparate data sources when fusing their data. Fusing two data sources, and specifically that of imaging and EHR (n=52) was the most prevalent as per Table 4. Figure 4 A showcases the relative distributions of preferred interface type and fusion strategy, the most popular being the Python platform and early fusion.

| Number of Modalities | Subtype | (# of papers) |
|---|---|---|
| 2 | Imaging/EHR | 52 |
|   | EHR/Text | 21 |
|   | Imaging/Genomic | 14 |
|   | Imaging/Time series | 5 |
|   | Imaging/Text | 4 |
|   | EHR/Genomic | 4 |
|   | EHR/Time Series | 3 |
|   | Genomic/Text | 1 |
| 3 | Genomic/Imaging/EHR | 13 |
|   | EHR/Imaging/Time series | 2 |
|   | Text/Imaging/EHR | 2 |
|   | EHR/Genomic/Text | 2 |
|   | Text/Imaging/Time series | 1 |
|   | EHR/Text/Time series | 1 |

*Table 4.* **Multimodalities and subtypes**

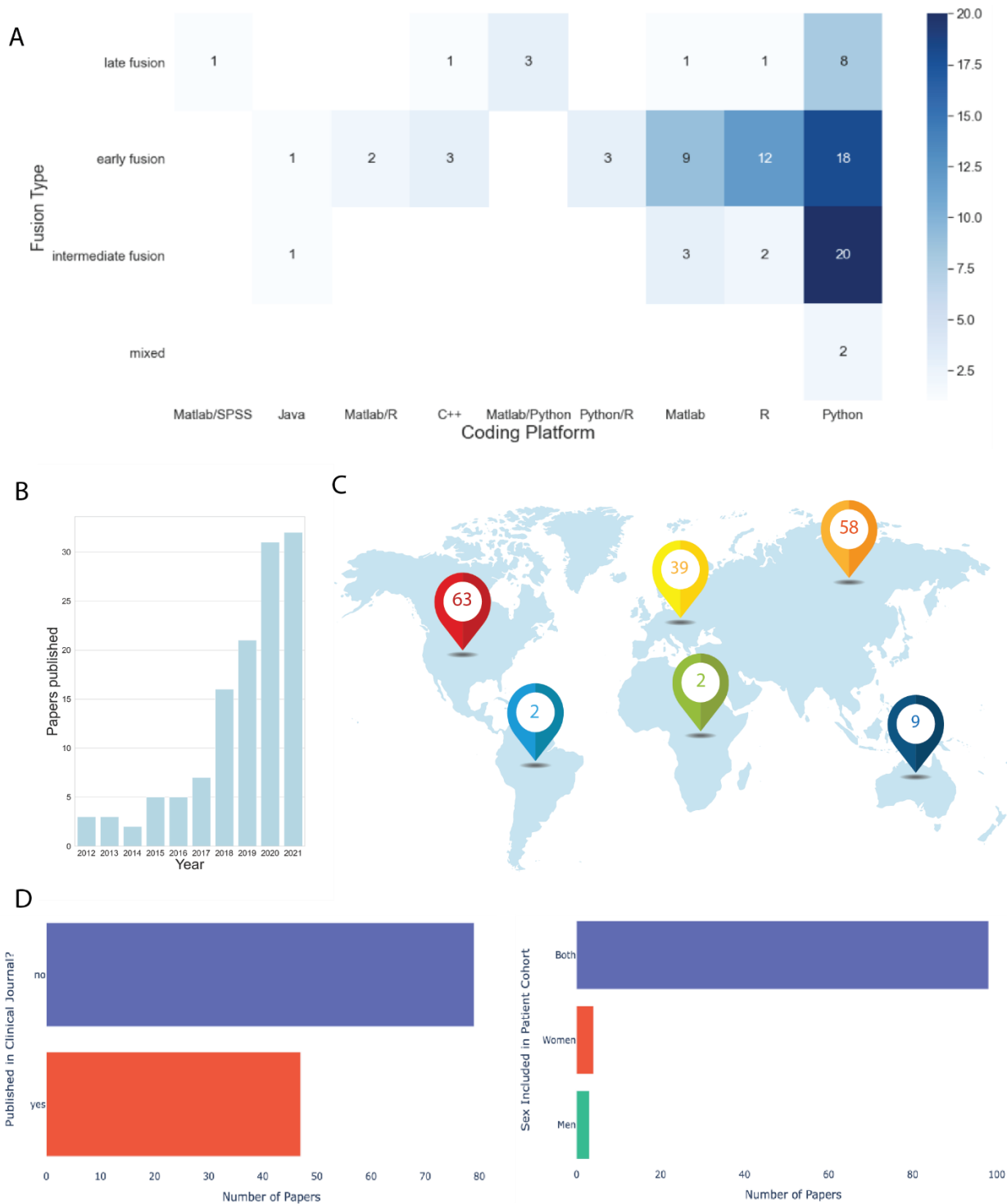

*Figure 4:* **A:** Heat map of fusion type broken down into the coding platforms papers reportedly used. **B:** Totally number of original research papers published in this sphere. **C:** Continental breakdown of author contributions (note some papers have authors from multiple continents). **D:** Breakdown of publication type (clinical/non-clinical journal) and sex breakdown of populations studied.

70 papers were published using early fusion. Of those, most were published using imaging and EHR data[31,33,46,47,52,55,56,59,61-63,67,70,72,74,84-87,89-92,94,100,106,107,110]. Nearly all these papers performed numericalization of their images prior to processing, however two performed matrix factorization[31,33].

A combination of EHR and text data was noted in 12 papers[28,53,68,71,78-80,95,98,102,108,109]. Meng et al. created a Bidirectional Representation Learning model used latent Dirichlet allocation (LDA) on clinical notes[98]. Cohen et al. used unigrams and bigrams in conjunction with medication usage[53]. Zeng et al. used concept identifiers from text as input features[80]. 9 papers used early fusion with imaging, EHR and genomic data[29,49,50,54,60,64,82,88,105]. Doan et al. concatenated components derived from images with polygenic risk scores[82]. Lin et al. also created aggregated scores from MRI, cerebral spinal fluid and genetic individually and brought them together into a single cohesive extreme learning machine to predict mild cognitive impairment[54]. Tremblay et al. used a multivariate adaptive regression spline (MARS) after normalizing, removing highly correlated features[88]. 10 papers performed fusion using imaging and genomic data[30,51,69,75-77,81,83,93,97]. Three of these generated correlation matrices as features by vectorizing imaging parameters and correlating them with single nucleotide polymorphisms (SNPs) prior to feeding into the model[30,69,77]. 3 papers in this category used EHR and time series[57,73,96]. Both Hernandez and Canni`ere et al. implemented their methods for purposes of cardiac rehabilitation and harnessed the power of support vector machines (SVMs). However, Hernandez preserved time series information by assembling ECG data into tensors that preserve the structural and temporal relationships inherent in the feature space[73], while Canni`ere performed dimensionality reduction of the time series information using t-SNE plots[57]. Two papers comprised early fusion using imaging and time series[66,103]. There were two papers that leveraged EHR and genomic information[65,140]. Luo et al. implemented hybrid non-negative matrix factorization (HNMF) to find coherence between phenotypes and genotypes in those suffering from hypertension[140]. A single paper performed early fusion using; (imaging and text data)[99] and (EHR, Genomics, Transcriptomics and Insurance Claims)[157].

33 papers were published using a form of intermediate data fusion. 14 used imaging and EHR data[58,111,113,116,119,121,123,125,127,130,131,133,135,137,138]. Zihni et al. merged the output from a Multilayer Perceptron (MLP) for modeling clinical data and Convolutional Neural Network (CNN) for modeling imaging data into a single full connected final layer to predict stroke[135]. A very similar approach was taken by Tang et al. who used 3-dimensional CNNs and merged the layers in the last layer[111].

EHR and text data were fused together in 11 papers[79,101,104,112,115,121,122,124,132,139,145]. Of these, six[79,112,122,124,132,145] used long term short term (LSTM) networks, CNNs or knowledge-guided CNNs[158] in their fusion of EHR and clinical notes. Chowdhury et al. used graph neural networks and autoencoders to learn meta-embeddings from structured lab test results and clinical notes[101,104]. Pivovarov et al. learned probabilistic phenotypes from clinical notes and medication/lab orders (EHR) data[139]. Two models each employing Latent Dirichlet Allocation (LDA) where data type is treated as a bag of elements and brings each and coherence between the two models outlines unique phenotypes. Ye et al. and Shin et al. used concept identifiers via NLP and bag-of-words techniques respectively prior to testing a multitude of secondary models[115,121]. In general, clinical notes can provide complementary information to structured

EHR data, where natural language processing (NLP) is often needed to extract such information[159-161].

A few were published using imaging and genomic[34,118,120]. Here radiogenomics are used to diagnose attention-deficit/hyperactivity disorder (ADHD), glioblastoma survival and dementia respectively. Polygenic risk scores are combined with MRI by Yoo et al. who used an ensemble of random forests for ADHD diagnosis[120]. Zhou et al. fused SNPs information together with MRI and positron emission tomography (PET) for dementia diagnosis by learning latent representations (i.e., high-level features) for each modality independently and subsequently learning joint latent feature representations for each pair of modality combination and finally learning the diagnostic labels by fusing the learned joint latent feature representations from the second stage[34]. Wijethilake also used MRI and gene expression profiling performing recursive feature elimination prior to merging into multiple models SVM, linear regression, artificial neural network (ANN) where the linear regression model outperformed the other two merged models and any single modality[118]. Wang et al. and Zhang et al. showcased their work in merging imaging and text information[128,129]. Both used LSTM for language modeling a CNN to generate embeddings that are joined together in a dual-attention model which is achieved by computing a context vector with attended information preserved for each modality resulting in joint learning. Seldomly were articles published in; (Imaging, EHR, Text)[114], (Genomic, Text)[48], (imaging, time series)[44], (Imaging, Text, Time series)[136], (Imaging, EHR, Genomic)[41], (Imaging, EHR, Time series)[117], (EHR, Genomic)[134], (EHR, text, time series)[126].

20 papers currently exist in this sphere having used late fusion. 7 of those performed this using imaging and EHR data types[43,141,142,149,150,155,162]. Both Xiong et al. and Yin et al. fed outputs into a convolutional neural network to provide a final weighting and decision[149,150]. 3 papers were published in using a trimodal approach: imaging, EHR and genomic[41,153,156].

Xu et al. and Faris et al. published papers using EHR and text data[45,152]. Faris et al. processed clinical notes using TF-IDF, hashing vectorizer and document embeddings in conjunction with binarized clinical data[45]. Logistic Regression (LR), Random Forest (RF), Stochastic Gradient Descent Classifier (SGD Classifier), and a Multilayer Perceptron (MLP) were applied to both sets of data independently and final outputs of the two models are combined using different schemes; ranking, summation, and multiplication. 2 articles were published using imaging and time series[148,151] both of which employed CNNs, one in video information of neonates[148] and the other in chest x-rays[151]. However, they differ in their processing of the time series data, where Salekin used a bidirectional CNN and Nishimori used a 1-dimensional CNN. Far fewer papers were published using (Imaging, EHR, Text)[42], (EHR, Genomic, Text)[147], (imaging, EHR, time series)[144], (imaging, genomic)[154], (EHR, Genomic)[143] and (imaging, Text)[36].

Two papers performed more than one kind of data fusion architecture[163,164]. Huang et al. created 7 different fusion architectures, these included early, joint and late fusion. The architecture that performed the best was the late elastic average fusion for the diagnosis of pulmonary embolism using computed tomography and EHR data[163]. Their Late Elastic Average Fusion leverages an ElasticNet (linear regression with combined L1 and L2 priors that act as regularizers) for EHR variables. El-Sappagh et al. performed early and late fusion to create an interpretable Alzheimer's diagnosis and progression detection model[164]. Their best performing model was one that implemented instance-based explanations of the random forest classifier by using the SHapley Additive exPlanations (SHAP) feature attribution. Despite using clinical, genomic and imaging data, the most influential feature was found to be the Mini-Mental State Examination.

Both men and women were represented relatively equally across papers, however varied in that distribution within an individual studies. Data fusion may help address sex differences and increase population diversity issues (including minority population) in health modeling by creating a more cohesive representative dataset if one datatype contained more of one and the reciprocal true for the other datatype(s). This would also hold true for racial or ethnic diversities. Less than half (37.6%) of the papers were published in a journal intended for a clinical audience. However, none of the papers included in the final cohort of studies had created tools for clinical use that had FDA approval. Figure 4 B contains the breakdown of total number of publications in this field over the last 10 years, and the Figure 4 C continental contributions of authors. Asia and North America had a near equal number of publications 60 and 61 respectively. Note that some titles had repeated counts due to cross-continent collaborations between authors.

**Discussion**

Returning to our research questions, we outlined from the inception of this work:
*RQ1- What characterizes the published literature using multi-modal data fusion in the health sector?*
The literature published in this area as displayed and characterized in the results' section is one that has a growing and global interest. It is fueled by a desire to improve predictive capabilities, relying on complimentary and correlative (reinforcing) data. The most common health topic was Neurology, and the second most was cancer. The relative saturation of this field likely reflects the underlying multimodal diagnosis that is present in a clinical counterpart i.e. neuroimaging and EHR (cognitive scores) as well as the number of established and curated databases that lends itself well for multi-modality predictions such as Alzheimer's Disease Neuroimaging Initiative[165] and The Cancer Genome Atlas Program[166]. Early data fusion methods leveraging imaging and EHR data likely owe their pervasiveness for reasons that are three-fold; 2 modalities over 3 is means less work overall in model building and deployment, EHR and image data do not require as extensive digital conversion for models such as text, and lastly early fusion is built on a single model with a multitude of feature inputs and is typically less computationally complex. Only 4 articles performed comparisons against their human clinician counterparts. Several did perform comparisons between uni-modal and multi-modal predictions, with the majority of those citing an improvement from baseline by incorporating heterogeneous data types. Of the studies that compared predictive performance between multi-modal data and uni-have modal data, there is a consistent citing of improvement in classification accuracy, sensitivity, and specificity. However, this was not seemingly limited to a particular subtype of multi-modal strategy that was detectable in our metadata. We therefore make the overall and general recommendation that multi-modal data integration be attempted to improve performance and better mirror a human expert by creating a higher validity environment from which to make clinical decisions.

*RQ2- What are the different analysis techniques, methods, and strategies applied to analyze multi-modal health data for diagnosis/prognosis?*

The analysis techniques are varied and currently do not showcase a gold standard or 'best-practice' in the field. This is likely linked to this being a relatively new and emerging field. Further, while an N-cross fold validation was the most common and a robust estimator in the face of bias within a dataset, strength of generalizability stems from either the dataset set containing multi-site/location patient data to begin with or using an external dataset from a remote location[167].

*RQ3*- *What areas of heterogeneous data fusion have had the most impact?*
Health contexts predominantly impacted by this include Neurology and Cancer predictive modeling, though no domain laid claim to building translation models via FDA (or equivalent) approval for use in clinical circumstances. As multi-modal fusion is touted to be an important fulcrum from which to leverage disparate types of information, there are calls to compare models more readily to physician decision makers[168-171]. This will guide the validity of the environments suitable to machine learning/artificial intelligence decision makers and hopefully result in the adoption and therefore FDA approval of these tools. In the review performed by Lyell et al.[172] they outline the methods of clinical decision support (assistive, autonomous information and autonomous decision), and acknowledge FDA approval of ML devices is a recent development. For without deployment, the benefits to these exercises remain minimal and indirect. Information fusion has led to an increase in predictive performances over single modalities[52,111,163].

The most common reported limitations in the papers included in the review are:
1) Cohort built on a single site/location: Samples were most often built from a single hospital or academic medical center[153]. A corollary of this issue is that most studies were based on samples from a single country. Because of the lack of an external cohort, it was difficult to validate and generalize the proposed model across different health systems.
2) Small Sample size: The median number of unique patients reported across the studies was 658 with a standard deviation of, 42600. This suggests that while some studies were able to leverage large and multi-center cohorts, a great many were not able to. Sample size were usually too small to train the model completely and obtain the best performance[69,81,120,125].
3) Retrospective data: Are Seldom machine learning investigations performed prospectively, this is endemic to the machine learning field in general[83].
4) Imbalanced samples: The imbalance problem in terms of positive and negative samples was usually ignored, which biases the model and is unfair to the validation test samples[74,150]. It is necessary to address the imbalance problem by adopting techniques such as under- or over-sampling or differential weighting for training samples.
5) Handling of missing data: Missing data were usually ignored by dropping data points or imputing. It is very important to address this issue, if not dealt with appropriately can skew the results due to a biased model[67,109,173]. More studies need to discuss frequencies and types of missing data, and the imputation method used, if any, to create a full data set[174-177]. Comparison of different imputation methods on the final results should be part of the reporting process[178].
6) Feature engineering: Extracting features completely are important to the predictive or classification model. The features extracted based on feature engineering may not be enough for training models, which is usually time-consuming and dependent on empirical knowledge from experts[110]. Deep learning-based techniques have been considered by researchers in some previous studies to extract features automatically from a dataset.
7) Confounding factors: When performing statistical analysis, researchers usually ignored

possible confounding factors such as age or gender, which may have major effects on the impact of results[42]. Such possible confounding effects should either be taken into consideration by the model[179,180] or adjusted first, prior to reporting model results.

8) Interpretation: Reasonable interpretations of the model and outputs must be presented so that clinicians find the results credible and then use them to provide guidance for treatments. However, most authors did not take the time to interpret the models for clinical audiences, and how the results provide useful tools. Different types of models need different types of explanation[41,123]. For example, using explainable tools such as Local Interpretation Model-Agnostic and Shapely Additive Explanation would clarify such feature contributions. Limitations are highlighted 'where' in the data processing and modeling building pipeline they exist, as per Figure 5.

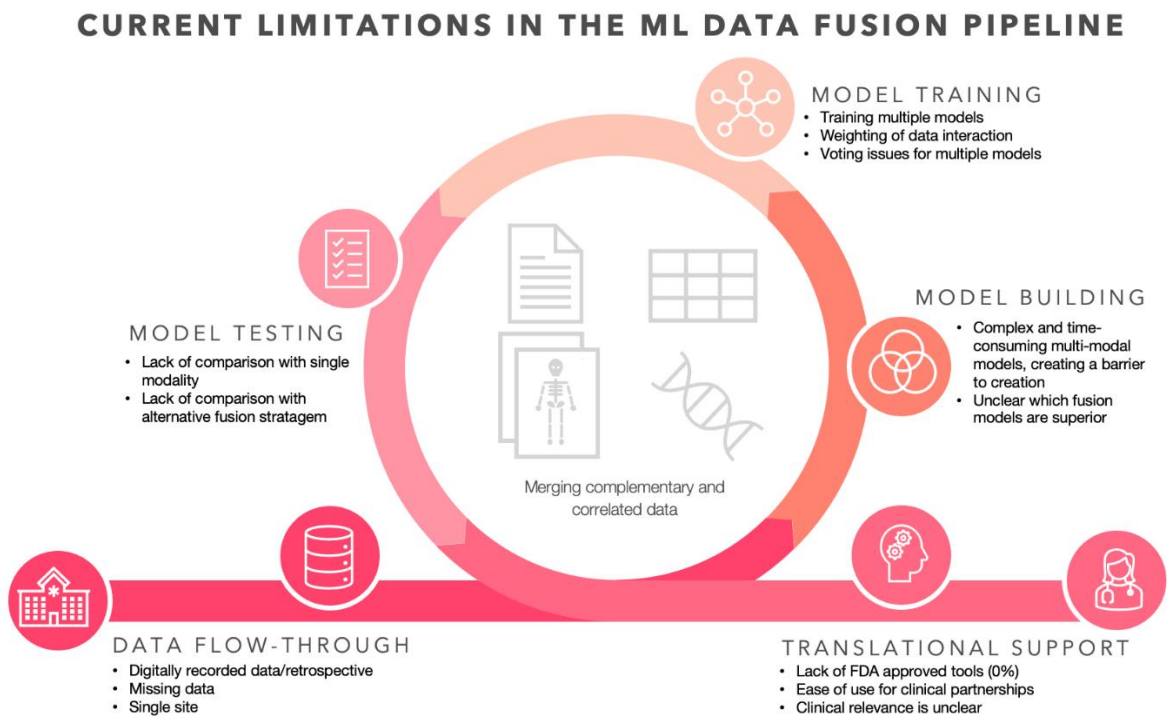

*Figure 5.* **Limitations to multimodal fusion in health**

To expedite and facilitate this field, we have outlined a number of gaps for future researchers in this field that we have garnered having performed this review. These are listed below, with specific suggestions.

The relative topic saturation showcased in Figure 3 above has highlighted notable places for future work. Two subdivisions that become apparent are; 1) several topics that would benefit from more work in that area, and 2) the need to progress health topics where significant effort has already been made. Medication/drug topics present an underrepresented area, with only two papers being published in this field[49,65]. Given current issues with drug interaction effects and recommendation guidelines such as the Beer's criteria[181] for use in older patients, performing multimodal machine learning may offer an earlier detection of medication misuse that is marked by iatrogenic error, user non-compliance or addiction.

Multimodal investigation may be a key to unlocking early detection, given the manifestations of misuse may be reflected in structured data (EHR) or imaging data. Take, for example, a heart failure patient who has inadvertently taken or been prescribed the wrong dosage of their diuretic[109]. Here this could be reflected in lab values (urine and serum) and systemic opacity on chest x-ray. The opioid crisis is wreaking havoc on the medical system, two hallmarks of which include pinpoint pupils and depressed respiratory status[182]. This also represents an area where image assessment of pupillary status and assessment of vitals (structured) may assist in making the decision to provide naloxone. Awareness of drug interaction effects is a difficult and growing issue[183-186], particularly in geriatrics, which gave rise to Beer's criteria. However, this does little to assess real time adverse events and may present another avenue of lucrative health returns if modeled with multimodal machine learning.

Yap et al. published the only dermatology geared paper contained in this review[47]. While some dermatological conditions are self-contained to the skin, many can be linked to gastrointestinal disease, infections or systemic cancer. Digital images of dermatological manifestations combined with lab values and clinical impressions via unstructured text presents a gateway of novel research questions. Therefore, the gauntlet has been thrown for pursuing a multimodal line of inquiry and exploration that could have both diagnostic and potentially screening implications in primary care settings. Similar justifications as outlined above could be applied to other areas seen as 'lacking' such as hematology with 1 paper[48] and nephrology having 3[86,108,112]. Conversely, the topics with more corresponding papers may improve their application and cement their gained expertise in our subsequent suggestions.

A multitude and extensive methodological undertaking is evident, characterized in the titles included in our review. While incorporating disparate data does lend itself to seemingly better predictions[142], our knowledge around certain diseases continue to accumulate and new data modalities continue to emerge. Thus, data fusion in healthcare is an evolving target that calls for machine learning framework to proactively adapt to the dynamic landscape[187].

It is anticipated this will vary based on the diagnosis or prognosis under investigation. For example, it has been shown in protein-protein interactions that utilization of the XGBoost algorithm reduces noisy features, maintain the significant raw features, and prevent overfitting via average gain[122]. XGBoost is a gradient boosting decision tree using regularized learning and cache-aware block structure tree learning for ensemble learning. The average gain is the total gain of all trees divided by the total number of splits for each feature. The higher the feature importance score of XGBoost is, the more important and effective the corresponding feature is[122]. Similarly, LightGBM is an ensemble algorithm developed by Microsoft that provides an efficient implementation of the gradient boosting algorithm[188]. LightGBM has the advantages of faster training speed, higher efficiency, lower memory usage, better accuracy, being capability for handling large-scale data, and the support of parallel and GPU learning[189], and has been consistently outperforming other models[190,191]. This represents a possible avenue for the advent of information fusion from disparate data in healthcare.

Augmenting clinical decision-making with ML to improve clinical research and outcomes present positive impacts that have economic, ethical, and moral ramifications as it has the ability to reduce suffering and save human lives. Multiple studies have pointed to that bias of the underlying data often lead to the bias in the resulting ML models[192,193]. The ethical imperative now on offer requires overcoming several hurdles with respect to data structure and federated access, clear definition of outcomes, assessment of biases and interpretability/transparency of results and limitations inherent in its predictions[194]. Future work is subsequently invited on a per-disease basis to characterize the combinatorics (which fusion strategy and set of mixed data types) optimize model performance. Doing so will push individual fields to create recommendations for subsequent real-world implementations.

Continuing in the same vein, model optimization with respect to disease-based predictions is the ease of use when modeling the data and the ensuing interpretability to end users. The extensive prepossessing and transformation inherent in multimodal machine learning create a significant and systemic drawback to its use. Perotte et al.[108] model was not compared with conventional simpler machine learning classifiers and their vectorization of clinical text required extensive manual work. And collective matrix factorization becomes inherently difficult to interpret[78]. Contrast this with Fraccaro et al. whose work into macular degeneration noted their white box performed as well as black box methods implementing[67].

Trade-offs between increasing accuracy at the expense of complex and time-consuming data transformations may mean the predictive power gained from a multimodal approach is offset by this front-end bottleneck and may mean predictions are no longer temporally relevant or useful. In addition to knowing how the models should be built per disease, we also advocate for pipelines and libraries to speed up data conversion processes via open access to make the technology more widely available[195,196].

Going from bench to bedside should be the goal of machine learning in medicine. Otherwise, these become technical exercises without real world impact. While lacking in the reviewed literature, it is understandable given the infancy of the machine learning health data fusion field. As machine learning and multimodal ML become more ubiquitous, there are increased demands for regulation and accountability.

Of crucial importance for uptake is that predictions be patient specific and actionable at a granular level[197]. For example, Golas et al. created a 30-day readmission prediction algorithm for those with heart failure[109]. If put into practice, it may serve to inform hospitals about resource management and prompt further lines of research that may decrease the number of patients being re-admitted in that 30 days. Linden et al. developed Deep personalized LOngitudinal convolutional RIsk model—DeepLORI (DeepLORI) capable of creating predictions that can be interpreted on the level of individual patients[122]. Simpler white-box models should be selected over complex ones if there exist similar performance measures to increase interpretability. Leveraging both and clinical and empirically driven information to create meaningful and usable recommendations[139] may improve clinician/end user under understanding by relating to existing frameworks. Resources such as CRISP-ML provide a framework for moving use cases into more practical applications[198], while efforts to vie for Food and drug administration (FDA) approvals as a tool for use are encouraged to increased adoption.

Limitations of this work include that it is not a systematic review. Therefore, it is possible that some titles that should have been included were missed. However, to mitigate this issue search strings were included based on both empirical knowledge in the field and alternative synonyms and keywords were identified using a literature review and language modelling and lexical analysis to find the context-sensitive terms that present the field. This was performed by using keyword co-occurrence strategy on each paper's objectives within the abstract of the article. As the primary purpose of this study was to perform scientific paper profiling on multimodal machine learning in health, a critical appraisal of individual methodological quality of the included studies was not performed. However, commentary is provided on the methodological limitations that could have affected their results and impact.

To our knowledge, this is the first review that offers a comprehensive meta-analysis evaluation across health domains, immaterial to the type of machine learning or the data used. This work serves as both a summary and steppingstone for future research in this field. The topic areas of health that have saturation relative to others were highlighted, which serves to build a foundation for future work. FDA approval and design of the tools for clinical centers also remains a large and outstanding invitation for further research to continue. Multi-site data integration is of the utmost importance for creating models that are generalizable and representative of the populace at large. Comparing developed models against working clinician counterparts would go a long way to effecting meaningful change in the uptake of algorithm-based decision-making models. Unimodal machine learning is inherently in contrast to current routine clinical practice in which imaging, clinical or genomic data are interpreted in unison to inform accurate diagnosis and warrants further work for ease of use and implementation. Overall, it appears justified to claim that multi-modal data fusion increases predictive performance over uni-modal approaches and is warranted where applicable and available results provide tenable models.

## Contributors
AK lead the review process, performed data extraction, performed the computation analysis, figure generation, writing and dashboard creation. HW, YL, SD, MH performed data extraction. ZX synthesized limitations of the studies. FW and FC performed proof reading and content curation. YL conceived the review, oversaw the review process and provided necessary feedback, proof reading, and content curation.

## Declaration of Interests
We declare no competing interests.

## Acknowledgments
The study is supported in part by NIH GrantsU01TR003528 and R01LM013337.